\pdfoutput=1
\documentclass{ieeeaccess}
\usepackage[table,xcdraw,dvipsnames]{xcolor} 
\usepackage{cite}
\usepackage{amsmath,amssymb,amsfonts} 
\usepackage{algorithmic}
\usepackage{graphicx}
\usepackage{textcomp}
\usepackage{bm}

\usepackage{amsmath,amsfonts,bm}

\def\eqref#1{equation~\ref{#1}}

\def\1{\bm{1}}

\DeclareMathAlphabet{\mathsfit}{\encodingdefault}{\sfdefault}{m}{sl}
\SetMathAlphabet{\mathsfit}{bold}{\encodingdefault}{\sfdefault}{bx}{n}

\usepackage{hyperref}
\usepackage{url}
\usepackage{booktabs}

\definecolor{lightgray}{gray}{0.9}
\definecolor{highlightblue}{RGB}{170, 220, 255} 
\usepackage{cleveref} 
\usepackage{subcaption}
      
\usepackage{pifont}         
\newcommand{\cmark}{\ding{51}}  
\newcommand{\tablehead}[1]{\textbf{\footnotesize #1}}

\makeatletter
\AtBeginDocument{\DeclareMathVersion{bold}
\SetSymbolFont{operators}{bold}{T1}{times}{b}{n}
\SetSymbolFont{NewLetters}{bold}{T1}{times}{b}{it}
\SetMathAlphabet{\mathrm}{bold}{T1}{times}{b}{n}
\SetMathAlphabet{\mathit}{bold}{T1}{times}{b}{it}
\SetMathAlphabet{\mathbf}{bold}{T1}{times}{b}{n}
\SetMathAlphabet{\mathtt}{bold}{OT1}{pcr}{b}{n}
\SetSymbolFont{symbols}{bold}{OMS}{cmsy}{b}{n}
\renewcommand\boldmath{\@nomath\boldmath\mathversion{bold}}}
\makeatother

\def\BibTeX{{\rm B\kern-.05em{\sc i\kern-.025em b}\kern-.08em
    T\kern-.1667em\lower.7ex\hbox{E}\kern-.125emX}}

\begin{document}

\history{Date of publication xxxx 00, 0000, date of current version xxxx 00, 0000.}
\doi{10.1109/ACCESS.2024.0429000}

\title{Deep Learning for Cross-Border Electricity Price Forecasting: A Comparative Study}

\author{Hadeer Elashhab*\authorrefmark{1},
Sai Srijan Papineni*\authorrefmark{2},
Marvin Dorn\authorrefmark{1}, 
Veit Hagenmeyer\authorrefmark{1}, 
and Benjamin Schäfer\authorrefmark{1}
}

\address[1]{Institute for Automation and Applied Informatics, Karlsruhe Institute of Technology, Karlsruhe, Germany (e-mails: \{hadeer.elashhab, marvin.dorn, veit.hagenmeyer, benjamin.schaefer\}@kit.edu)}
\address[2]{Department of Computer Science, Karlsruhe Institute of Technology, Karlsruhe, Germany (e-mail: saisrijanp@gmail.com)}

\tfootnote{The symbol * denotes that these authors contributed equally to this work.}

\markboth
{Elashhab et al.: Deep Learning for Cross-Border Electricity Price Forecasting: A Comparative Study}
{Elashhab et al.: Deep Learning for Cross-Border Electricity Price Forecasting: A Comparative Study}

\corresp{Corresponding author: Hadeer Elashhab (e-mail: hadeer.elashhab@kit.edu).}

\begin{abstract}
While publicly available electricity market data presents a valuable resource for forecasting research, the field lacks established benchmark datasets for standardized comparison. As a result, many studies have relied on different datasets and metrics to evaluate methods in isolated settings, making it difficult to assess progress and compare state-of-the-art approaches consistently.
In this work, we use public data to evaluate deep learning models for electricity price forecasting (EPF) across multiple market settings. Our goal is to establish a reproducible framework that enables a consistent evaluation of forecasting models.
Developing standardized benchmarks for EPF is particularly important given the growing complexity of electricity markets, driven by the increasing integration of renewable energy sources. Their volatility increases the supply uncertainty and creates additional forecasting challenges. Under these conditions, accurate EPF methods support operational efficiency, energy trading, and grid stability.
Although deep learning has been explored for day-ahead EPF, many prior studies are limited to single-market settings, narrow feature sets, or fixed training regimes. This work presents a comparative evaluation of six deep learning models--covering state-space, MLP, RNN, and Transformer-based architectures--emphasizing generalization across markets.
We simulate low-data target-market conditions using zero-shot, one-shot, and few-shot learning. Our test set focuses on the Germany–Luxembourg (DE-LU) bidding zone in 2024 using a standardized dataset with calendar, historical price, and market-derived features.
Our findings suggest that N-HiTS and NBEATSx perform competitively in limited-data scenarios, while transformer-based models can reach comparable accuracy but tend to require more adaptation and tuning. Model performance also benefits from careful feature selection and hyperparameter tuning, and we note that the differences between the strongest models are often small.
\end{abstract}


\titlepgskip=-21pt
\maketitle

\section{Introduction}
\label{sec:intro}
Addressing climate change has motivated a global shift toward sustainable energy systems. A key element of this transition is the integration of renewable energy sources, such as wind, solar, and hydro, which can help reduce greenhouse gas emissions and reliance on fossil fuels. In addition to their environmental advantages, renewables are becoming increasingly competitive with conventional energy sources \cite{nations_renewable_nodate}.

Electricity price forecasting (EPF) plays a critical role in supporting the energy transition by improving decision-making in electricity markets. While the integration of weather-dependent energy sources, such as solar and wind, relies heavily on accurate weather forecasts, EPF provides a complementary and often more actionable signal. Market prices implicitly capture a wide range of system conditions, including supply-demand imbalances, generation costs, and grid constraints, without requiring explicit communication of generation capacities, maintenance schedules, or outages. As such, price signals offer a compact and observable indicator of system stress or surplus. Accurate price forecasts can inform demand-side management, guide the dimensioning of backup generation, and support charging strategies for energy storage systems such as batteries. For market participants, EPF enables more effective bidding strategies in the day-ahead and real-time markets. Similarly, for consumers and system operators, EPF facilitates cost management, operational planning, and system reliability \cite{kuo_electricity_2018,cramer_multivariate_2023}.

While deep learning has shown potential for improving EPF, much of the literature focuses on isolated market contexts, short evaluation windows, and single-model assessments. To complement this, we present a comparative analysis of six deep learning models, covering different architectures--Neural Hierarchical Interpolation for Time Series (N-HiTS) \cite{challu_NHITS_2023}, Neural Basis Expansion Analysis with Exogenous Variables (NBEATSx) \cite{oreshkin_n-beats_2020,olivares_neural_2023}, Temporal Fusion Transformers (TFT) \cite{lim_temporal_2021}, Mamba \cite{gu_mamba_2024}, Long Short-Term Memory (LSTM) \cite{hochreiter_long_1997}, and Vanilla Transformer (VT) \cite{vaswani_attention_2017}--in a cross-border setting. Our analysis focuses on day-ahead EPF for the Germany–Luxembourg (DE-LU) bidding zone, which is among the most interconnected and economically significant regions in Europe.

Our evaluation covers the full year 2024, a period characterized by dynamic market conditions, allowing us to test model robustness in a realistic and challenging context. To simulate practical constraints in cross-border forecasting scenarios, we adopt three learning settings: zero-shot (no market data from DE-LU in the training set), one-shot (one sample of DE-LU market data in the training set), and few-shot learning (a few samples of DE-LU market data in the training set). We also explore the effects of input feature selection, temporal alignment, and training window size on forecasting performance.

The contributions of this study are as follows: (1) a comparative study of six deep learning models under multiple data regimes, (2) a reproducible forecasting pipeline based on publicly available market data, and (3) the development of a benchmark that leverages this underutilized dataset to support future work on data-efficient and generalizable EPF. We aim to encourage the broader use of this dataset in benchmarking studies by providing a transparent evaluation framework and detailed experimental setup.

The remainder of this paper is organized as follows: \Cref{sec:rel} reviews related work, \Cref{sec:data} describes the dataset and preprocessing steps, \Cref{sec:methods} outlines the benchmarking methodology, \Cref{sec:res} presents the experimental results, and \Cref{sec:limitations} provides a discussion of the limitations. Finally, \Cref{sec:sum} presents conclusions and outlines directions for future research.

\section{Related Work}
\label{sec:rel}

EPF has been a long-standing area of research within energy systems, with early efforts relying primarily on classical statistical methods such as autoregressive integrated moving average (ARIMA), generalized autoregressive conditional heteroskedasticity (GARCH), and vector autoregression (VAR) models. For instance, some researchers compare ARIMA and GARCH across six European countries, showing their ability to model seasonality and volatility \cite{tehrani_electricity_2022}. Similarly, another study evaluated ARIMA(X), convolutional neural network--long short-term memory (CNN-LSTM) hybrids, and VAR models and found that while statistical approaches can remain effective for short-horizon predictions, they tend to be less robust in highly dynamic or nonlinear environments \cite{lehna_forecasting_2022}.

More recently, deep learning methods have been applied to EPF, with a focus on capturing complex temporal patterns and nonlinear relationships. For example, one group proposed a hybrid convolutional neural network--gated recurrent unit (CNN-GRU) model \cite{lehna_forecasting_2022}, and  another introduced a system combining copula-based feature selection, signal decomposition, and Bayesian optimization and hyperband (BOHB)-tuned LSTM \cite{xiong_hybrid_2023}. These models report improved accuracy in specific market contexts, although their architectures often require extensive feature engineering and careful hyperparameter optimization. Notably, these evaluations are typically conducted in single-market settings.

Lago et al. provide a broad comparison of traditional and machine learning models for EPF \cite{lago_forecasting_2021}. Their open-access framework emphasizes the importance of reproducibility, consistent baselines, and statistical rigor in empirical evaluations. While their work sets a foundation for standardized comparisons, it does not extensively explore cross-border learning or generalization across different markets.

Research on spatial generalization has begun to emerge in response to increasing interconnections within European electricity markets. Some studies incorporate cross-border price features and use SHapley additive explanations (SHAP) values for interpretability \cite{tschora_electricity_2022}, while another study demonstrates that including features from interconnected markets can improve forecasts for the Dutch market \cite{heijden_electricity_2021}. These studies highlight the relevance of spatial coupling but often focus on fixed training regimes rather than transfer learning or low-data scenarios.

Recent developments have focused on the integration of exogenous inputs. For example, one study extends the N-BEATS architecture to NBEATSx, thereby enabling the inclusion of various static and temporal features \cite{olivares_neural_2023}. They report improved performance relative to classical and deep learning baselines. In a broader analysis, another study evaluates deep multilayer perceptron (MLP)-based models across 19 European bidding zones and shows that market volatility is not always predictive of model performance \cite{aliyon_deep_2024}. However, both studies emphasize supervised settings with moderate-to-large training data availability.

The question of generalization to unseen markets has motivated an interest in transfer learning. This has led some researchers to explore pre-training on source markets and fine-tuning on target markets such as France and Germany \cite{gunduz_transfer_2023}. Although this approach demonstrates gains in specific settings, it does not consider strategies such as zero-shot or few-shot learning, which may be more reflective of real-world data availability constraints. Our work complements this direction by explicitly comparing multiple learning strategies under low-data and cross-border conditions.

In parallel, probabilistic methods are gaining traction in EPF owing to their potential for risk-aware decision-making. One study applies distributional neural networks to forecast German prices and shows that probabilistic outputs could improve downstream trading performance \cite{marcjasz_distributional_2023}. Another study  leverages normalizing flows for multivariate intraday forecasting, emphasizing uncertainty quantification and explainability \cite{cramer_multivariate_2023}. These studies advance methodological tools but do not benchmark generalization strategies or cross-market transferability.

In summary, although deep learning has enabled substantial progress in EPF, many existing studies are confined to isolated markets, fixed data regimes, or single-model evaluations. Our work aims to contribute to the growing literature by evaluating multiple deep learning architectures within a unified framework, with a focus on data-efficient transfer learning across markets. We also provide a reproducible pipeline using publicly available data to facilitate future comparative studies of cross-border forecasting.

\section{Data}
\label{sec:data}

The electricity price data used in this study is sourced from the Energy Charts platform \cite{burger_energy-charts_nodate} and its associated API \cite{noauthor_energy-charts_nodate}, both maintained by the Fraunhofer Institute for Solar Energy Systems (Fraunhofer ISE). The license and terms of use can be found in \Cref{sec:license}, which provides access to day-ahead hourly spot market prices for European bidding zones in EUR/MWh with consistent coverage beginning in 2015.

\begin{figure}[hbt!]
    \centering
    \includegraphics[width=0.6\linewidth]{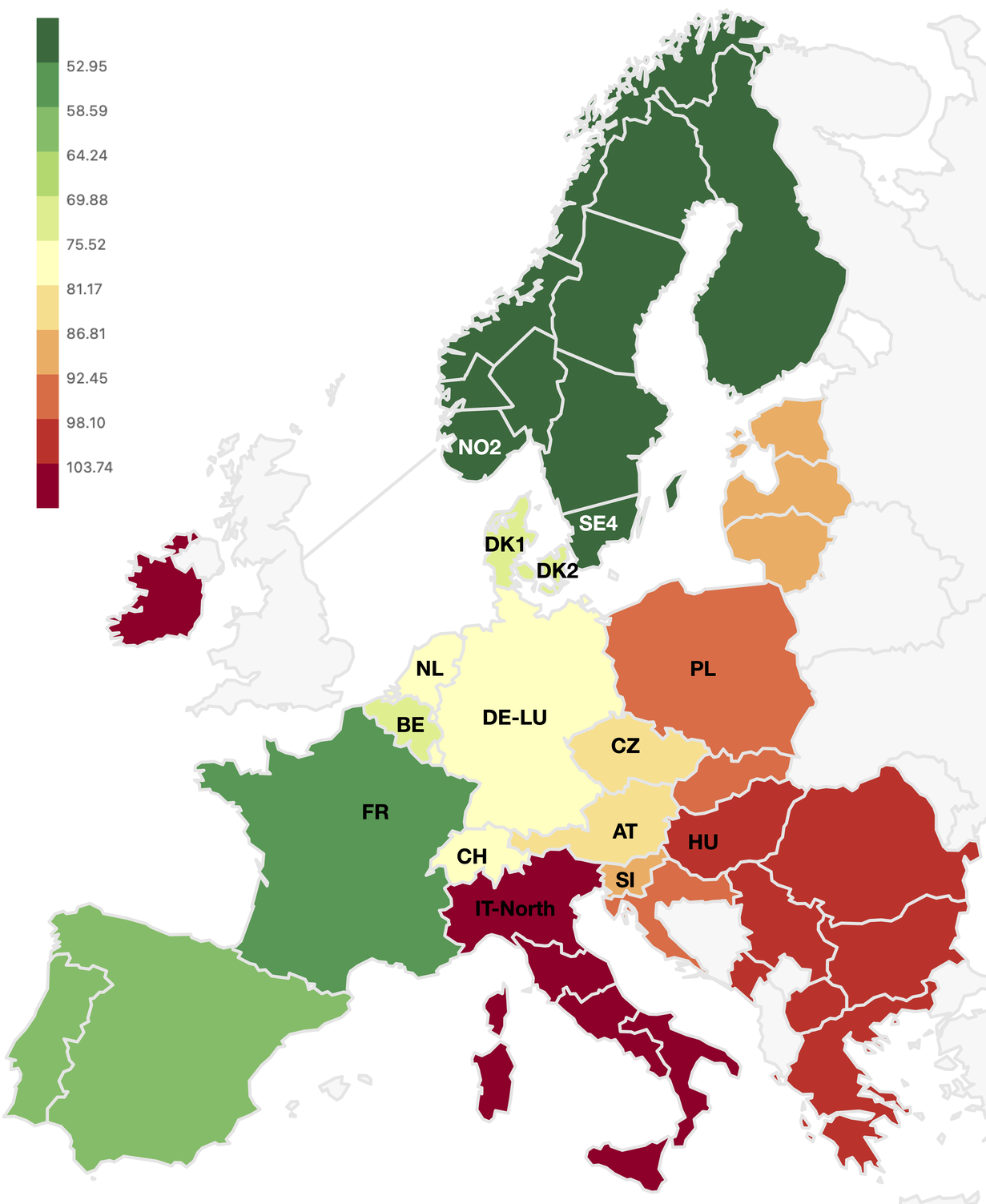}
    \caption{Average electricity spot market prices across European bidding zones in 2024 (EUR/MWh). The color indicates average annual price levels, with dark red representing higher prices and dark green representing lower prices. Bidding zones labeled with their respective codes (e.g., DE-LU, FR, DK1) are those included in the learning strategy experiments presented in this work. Adapted from \cite{noauthor_borsenstrompreise_nodate} with permission.}
    \label{fig:bidding-zones-map}
\end{figure}

In the European electricity market, bidding zones are defined as the geographical areas within which a single market-clearing electricity price is established. These zones typically align with national borders; however, exceptions exist where a single country may have multiple zones, such as Italy, or where two countries share a zone, such as Germany and Luxembourg (see \Cref{fig:bidding-zones-map}).

Another definition could be that a bidding zone is a geographical area within the electricity market where electricity can be bought and sold without considering physical grid limitations \cite{noauthor_bidding_nodate}.

While historical data for the former Germany–Austria– Luxembourg (DE-AT-LU) bidding zone is available from as early as 2006, we restrict our dataset to the period starting in the last quarter of 2018 to reflect the structural market change that occurred when this joint bidding zone was split into two separate zones, DE-LU (Germany and Luxembourg) and AT (Austria). This temporal constraint ensures consistency in the spatial definition of price zones throughout the analysis. However, when incorporating cross-border data, we include data starting from 2015 but explicitly exclude all data from Austria to avoid potential inconsistencies.

The primary dataset includes day-ahead electricity prices (EUR/MWh) along with exogenous variables that may influence price formation. These include calendar features, market factors, electricity load, cross-border trading data, and energy generation shares, providing a broad representation of key market dynamics.

\subsection*{Non-price features}

In addition to electricity prices, our dataset incorporates a range of exogenous features to capture temporal patterns and relevant market signals. Calendar features are constructed using cyclic and one-hot encoding to reflect the inherent periodicity in electricity demand and pricing. Specifically, we encode the day of the week, month, and hour using sine and cosine transformations (\texttt{day\_of\_week\_sin}, \texttt{day\_of\_week\_cos}, etc.) to preserve the cyclical structure. Binary indicators are included to mark weekends and public holidays based on the official German calendar, as these days typically exhibit distinct load and pricing behaviors.

Market variables include fundamental drivers of electricity price formation in Germany. These consist of natural gas prices, the synthetic price benchmark, CO\textsubscript{2} emission allowance prices, and electricity load data. These features collectively reflect aspects of supply-demand dynamics, regulatory costs, and broader economic influences relevant to price forecasting.

The synthetic price is a feature-engineered variable that combines gas prices and CO\textsubscript{2} emission allowances (\Cref{eq:synthetic_price_formula}) to approximate the electricity price formation. The formula is adopted from work by Leonhard Probst \cite{fraunhofer_ise_energy_2025}.
This feature provides an estimation of the average spot prices in the Day-Ahead Auction (volume-weighted for the DE-LU Zone) (\Cref{fig:synthetic_price}), offering insights into market dynamics and price-setting mechanisms. The results indicate that electricity prices are highly dependent on gas prices and CO\textsubscript{2} emission allowances, highlighting their influence on market fluctuations. The dependency of electricity prices on gas prices arises from the merit order principle, because the market price is typically set by the most expensive power plant needed to meet demand, often a gas-fired plant.

The efficiency of modern combined-cycle gas turbine (CCGT) plants is typically above 55\% today \cite{noauthor_sgt5-8000h_nodate}. We adopt 55\% as a representative value, at which the synthetic price aligns well with observed day-ahead prices. Remaining deviations are likely attributable to producer profit margins, additional operating costs, and the presence of older, less efficient plants in the generation mix.

\begin{equation}
  P_{\text{syn}} = \frac{P_{\text{gas}}}{\eta} + \varepsilon_{\text{CO}_2} \cdot P_{\text{CO}_2}
  \label{eq:synthetic_price_formula}
  \end{equation}

  \noindent where
  \begin{IEEEdescription}[\IEEEsetlabelwidth{$\varepsilon_{\text{CO}_2}$}]
    \item[$P_{\text{syn}}$] synthetic electricity price (EUR/MWh);
    \item[$P_{\text{gas}}$] natural gas price (EUR/MWh);
    \item[$P_{\text{CO}_2}$] CO\textsubscript{2} certificate price (EUR/tCO\textsubscript{2});
    \item[$\eta$] plant efficiency, set to $55\%$;
    \item[$\varepsilon_{\text{CO}_2}$] emission intensity, set to $0.4$\,tCO\textsubscript{2}/MWh.
  \end{IEEEdescription}

\subsection*{Data known at forecasting time}

In the DE-LU bidding zone, day-ahead electricity prices are determined through uniform-price auctions conducted by the European Power Exchange (EPEX SPOT). These auctions occur daily at 12:00 CET, and the resulting prices apply to each hourly interval of the following day (00:00–23:00 CET). To prepare forecasts ahead of these auctions, the models rely on data available prior to noon on the day before delivery. While some exogenous features, such as calendar data, gas prices, CO\textsubscript{2} emission allowances, and cross-border trading volumes are available in advance, others, such as load and generation data, are typically reported only up to two hours before the current time.

To address this, we adopt a Full Previous Day as Input strategy, using the latest 24 hours of available data and applying imputation where needed. We evaluated several imputation techniques and found that weekly historical values produced stable estimates for load, whereas previous-day values were more appropriate for renewable and non-renewable generation. This approach is intended to approximate realistic market conditions while maintaining a consistent input quality for forecasting.

\section{Framework}
\label{sec:methods}

\subsection{Training and Test Sets}

To evaluate the generalization capacity of our models across different historical regimes, we train them on three datasets of varying lengths. The largest spans from October 2018 to the end of 2023, covering relatively stable periods, the COVID-19 shock, and the 2022 energy crisis \cite{goldthau_energy_2022,ozili_global_2023,trebbien_patterns_2024}.
A medium-sized dataset (2020–2023) emphasizes pandemic-driven market disruptions and recovery, balancing the historical context and model complexity \cite{ozili_global_2023,narajewski_changes_2020}. The smallest dataset, limited to 2023, focuses on recent volatility, and may be more suitable for simpler models. These splits reflect major structural changes and facilitate testing under various market dynamics.

\begin{figure}[h]
    \centering
    \includegraphics[width=1\linewidth]{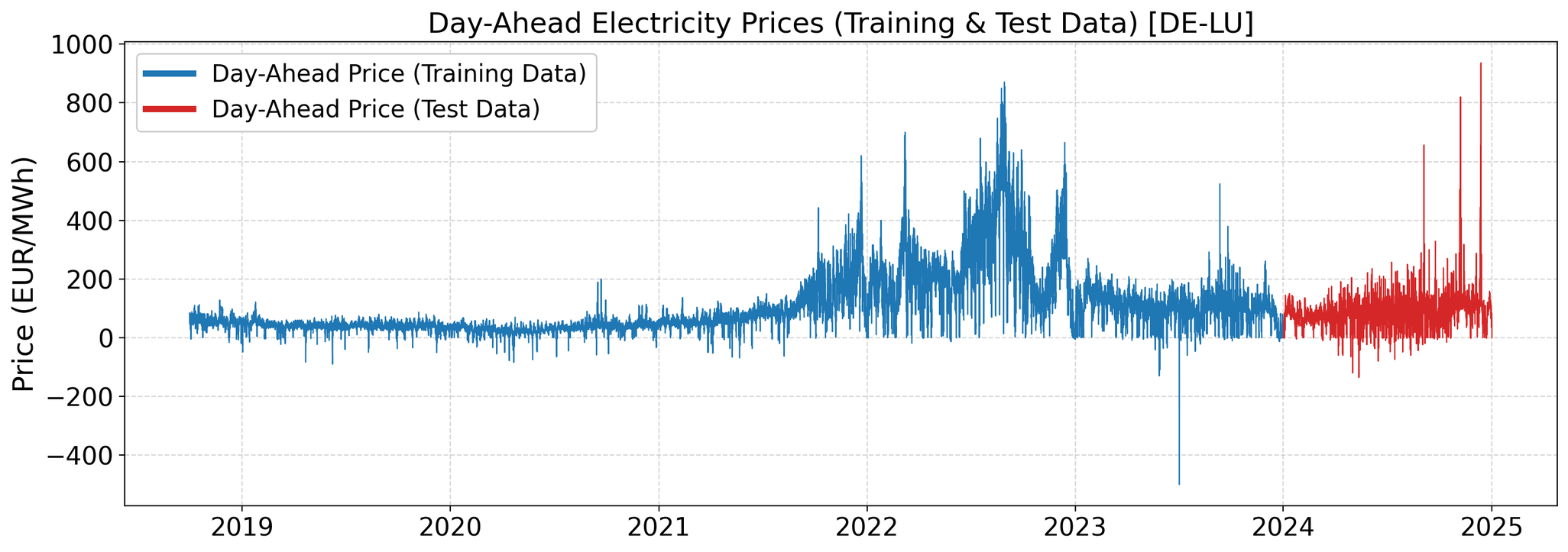}
    \caption{Day-ahead electricity prices (EUR/MWh) for the DE-LU bidding zone. The training data (blue) spans from 2018-10-01 to 2023-12-31, while the test data (red) contains prices for 2024. The figure highlights extreme price spikes, particularly during the 2022 energy crisis, where surging gas prices had a direct impact on electricity costs. Notably, the highest recorded prices are observed in the test dataset, reflecting the day-ahead electricity spot market prices of 2024.}
    \label{fig:training_data_large}
\end{figure}

The test dataset spans the full year 2024 and consists exclusively of electricity prices from the DE-LU bidding zone. This dataset serves as a consistent evaluation set across all experiments, including zero-shot, one-shot, and few-shot learning scenarios (see \Cref{fig:training_data_large}).

We study the DE-LU bidding zone in detail for two primary reasons. First, it is the largest bidding zone in Europe, encompassing all of Germany and serving over 80 million people. Second, it is highly interconnected with several other bidding zones, including continental European and Nordic zones \cite{trebbien_patterns_2024,noauthor_bidding_nodate-1}.

The choice of 2024 allows us to examine the model behavior under current and dynamic market conditions, characterized by price volatility, policy shifts, and increasing renewable integration. Each quarter includes different types of challenges, ranging from price spikes to periods of relative stability, offering a diverse testbed for assessing forecasting robustness and generalization.

For model training, we use overlapping sliding windows with an hourly stride primarily used to increase the number of training samples and help the models learn fine-grained temporal dependencies. In some cases, hyperparameter optimization may select a daily stride resulting in non-overlapping windows that reduce redundancy but may provide less exposure to short-term fluctuations.

\subsection{Model Selection and Comparison}

We compare six deep learning models for day-ahead EPF (DAEPF), covering a range of deep learning architectures. These include: 

\begin{itemize}
\item LSTM:
LSTM is a recurrent neural network (RNN) architecture designed to overcome the vanishing gradient problem. While standard RNNs fail to learn long-range dependencies due to decaying error signals, LSTM maintains a stable gradient flow over extended time intervals.
The architecture centers on memory cells containing constant error carrousels (CECs), which preserve information across thousands of time steps. Access to these cells is regulated by multiplicative gate units: the input gate protects the cell from irrelevant signals, while the output gate controls when stored information is released to the rest of the network. This mechanism allows the model to efficiently bridge long time lags with $O(1)$ computational complexity per time step.
\item NBEATSx:
Building upon the N-BEATS framework \cite{oreshkin_n-beats_2020}, NBEATSx extends the original architecture to address the critical limitations of covariate integration and model interpretability. While the original N-BEATS demonstrated state-of-the-art performance using a deep stack of fully-connected layers and residual links, it was restricted to univariate data. NBEATSx introduces a specialized substructure utilizing convolutional layers to encode both static and time-dependent exogenous variables while preserving temporal dependencies. Furthermore, the architecture enhances interpretability by enabling a comprehensive signal decomposition; it can isolate the effects of level, trend, and seasonality, alongside the specific impacts of exogenous variables. 
\item N-HiTS:
The N-HiTS architecture is an evolution of the N-BEATS framework designed specifically to address the challenges of long-horizon forecasting. While maintaining the core structure of residual learning through stacks of multilayer perceptron (MLP) blocks, N-HiTS introduces a hierarchical approach to signal processing that significantly reduces computational complexity while improving accuracy. 
The model operates on two primary novel principles: 1) Multi-Rate Data Sampling, where each block employs sub-sampling layers to capture a broad range of temporal dependencies--from high-frequency fluctuations to long-term trends--without the memory overhead of processing the full-resolution signal; and 2) Multi-Scale Synthesis, which utilizes hierarchical interpolation to ensure smooth long-horizon predictions by generating low-dimensional coefficients that are subsequently upsampled. By synchronizing the input sampling rate with the output interpolation scale, each block specializes in a specific frequency band, allowing N-HiTS to outperform Transformer-based techniques with an order-of-magnitude reduction in compute costs.
\item Mamba:
Mamba is a linear-time sequence modeling architecture that addresses the computational inefficiencies of Transformers while maintaining comparable modeling power. At its core, Mamba introduces a selection mechanism that allows structured state space models (SSMs) to be input-dependent, enabling the model to selectively propagate or forget information along the sequence dimension. This "content-aware" reasoning overcomes the fundamental limitations of prior linear time-invariant (LTI) models, which struggled with discrete and information-dense data.
To maintain efficiency despite the time-varying nature of its parameters, Mamba utilizes a hardware-aware parallel algorithm. This approach computes the model recurrently using a scan instead of a convolution, leveraging a fused kernel to exploit the GPU memory hierarchy and avoid materializing the full latent state in slow high-bandwidth memory. Structurally, the architecture simplifies deep sequence models by integrating selective SSMs into a homogeneous block design that omits standard attention and MLP layers. Consequently, Mamba achieves linear scaling in sequence length and provides up to five times higher inference throughput than Transformers, while reaching state-of-the-art performance across language, genomics, and audio modalities.
\item VT:
The VT represents a fundamental shift in sequence modeling by entirely dispensing with recurrence and convolutions in favor of a pure attention-based architecture. The model follows an encoder-decoder structure where both components are composed of a stack of identical layers--typically six in the base configuration. The core innovation is the multi-head self-attention mechanism, which allows the model to jointly attend to information from different representation subspaces at different positions, effectively capturing global dependencies regardless of their distance in the sequence.
In terms of internal architecture, each encoder layer consists of two sub-layers: a multi-head self-attention mechanism and a position-wise fully connected feed-forward network. The decoder adds a third sub-layer to perform attention over the encoder's output, and its self-attention is masked to preserve the auto-regressive property by preventing positions from attending to subsequent tokens. To facilitate training in deep stacks, the model utilizes residual connections around each sub-layer followed by layer normalization. Because the architecture lacks inherent sequential awareness, positional encodings--specifically sinusoidal functions--are added to the input embeddings to inject information about the relative and absolute order of tokens. This design enables significantly more parallelization during training compared to recurrent models.
\item TFT:
The TFT is an attention-based architecture designed for high-performance multi-horizon time series forecasting that prioritizes interpretability alongside accuracy. To handle the complexity of real-world datasets, the model integrates specialized components such as Variable Selection Networks to identify salient features and Gated Residual Networks to adaptively suppress irrelevant inputs. 
Its temporal processing is divided into a decoder that uses recurrent LSTM layers for local context and interpretable multi-head self-attention layers to capture long-term dependencies. Unlike standard black-box models, the TFT explicitly incorporates static metadata, observed historical data, and known future inputs, while producing quantile forecasts to represent prediction uncertainty across various time scales.
\end{itemize}

This selection spans different modeling strategies, which enables a comparative analysis of their performance with respect to temporal dependencies, exogenous input handling, and longer-range forecasting.

All models except Mamba are implemented using the NeuralForecast library, which standardizes the training and inference procedures across architectures \cite{olivares2022library_neuralforecast}. Mamba was integrated via its official implementation and extended to support historical exogenous inputs for the DAEPF \cite{noauthor_state-spacesmamba_2025}. Each model supports different combinations of exogenous variables, and their computational complexity ranges from linear (Mamba) to quadratic (transformer).

To align with common practices in the EPF literature, we evaluate the point forecast accuracy using the mean absolute error (MAE) and root mean squared error (RMSE). While regression-based models are sometimes optimized via squared error, we train our models using absolute error to remain consistent with the broader literature.

\subsection{Input Features}

To promote comparability across architectures, all models are trained using a shared set of input features grouped into three categories:
\begin{itemize}
    \item \textbf{Calendar features:} always available and include hour-of-day, day-of-week, month, weekends, and German public holidays.
    \item \textbf{Future-known market variables:} include gas prices, CO\textsubscript{2} emission allowances, and a derived synthetic price.
    \item \textbf{Historically reported features:} include electricity load, renewable generation, and non-renewable generation; these are typically available up to 2 hours before the day-ahead auction (announced daily at 12{:}00~CET).
\end{itemize}

To evaluate generalization independently of exogenous market data, we limit the input features to calendar variables in zero-, one-, and few-shot experiments. All available exogenous features are used for full-model training and tuning, unless otherwise noted.

\textbf{Time Representation and Preprocessing}

Proper temporal alignment is important in DAEPF. Feature extraction is performed in local time (CET) to preserve the correct calendar relationships and holiday effects. Timestamps are then converted to UTC before training to ensure continuity and avoid duplication or gaps around daylight saving time transitions. This dual-step preprocessing helps to maintain consistency across time and supports interpretability.

\subsection{Hyperparameter Optimization}

All models are tuned using the AutoModel framework by NeuralForecast together with Optuna’s tree-structured parzen estimator (TPE) for Bayesian optimization \cite{akiba_optuna_2019}. This setup enables the systematic exploration of the hyperparameter space based on validation feedback.

All six models are tuned using the full set of input features, as described previously. Search spaces are initialized based on default recommendations and extended to cover DAEPF-specific settings. We use a fixed 90/10 train-validation split or cross-validation (6–12 folds) depending on the model type and dataset size. Lighter models (e.g., NBEATSx and N-HiTS) are tuned with up to 30 trials, whereas more computationally intensive models (e.g., TFT and VT) are limited to 10 trials. The Adam optimizer is used across models because of its reliability in non-stationary environments.

\subsection{Learning Strategies}

To assess the generalization under different levels of data availability, we evaluate three learning strategies with DE-LU as the target zone:
\begin{itemize}
    \item \textbf{Zero-Shot Learning (ZSL):} No DE-LU data is seen during training.
    \item \textbf{One-Shot Learning (OSL):} A single training sample from DE-LU, consisting of a 7-day (one-week) input window followed by a 1-day (24h) forecast horizon as the output.
    \item \textbf{Few-Shot Learning (FSL):} This includes 30 consecutive days of DE-LU.
\end{itemize}

These settings are intended to simulate practical forecasting scenarios, in which the amount of target-market data may be limited. \Cref{tab:learning_strategies} summarizes the data availability for each strategy.
\begin{table}[hbt!]
    \caption{Overview of DE-LU exposure under each learning strategy.}
    \centering
    \resizebox{\columnwidth}{!}{%
    \begin{tabular}{lll}
        \toprule
        \textbf{Learning Strategy} & \textbf{DE-LU Training Data} & \textbf{Purpose} \\
        \midrule
        Zero-Shot (ZSL) & 0 days & Full generalization \\
        One-Shot (OSL) & 7 days + 1 forecast window & Minimal adaptation \\
        Few-Shot (FSL) & 4 weeks & Partial adaptation \\
        \bottomrule
    \end{tabular}
    }
    \label{tab:learning_strategies}
\end{table}

\section{Results}
\label{sec:res}
\subsection{Models}
\subsubsection{Baseline Models}
\label{sec:baseline_models}

Before introducing the deep learning models, we evaluate simple baselines based on historical price patterns and basic market factors. These methods serve as intuitive reference points for assessing the added value of complex forecasting models.

\begin{itemize}
    \item \textbf{Synthetic Price:} A formula-based estimate derived from gas prices and CO\textsubscript{2} emission certificates, as defined in \Cref{eq:synthetic_price_formula}.
    \item \textbf{Previous Day’s Price:} Assumes that the electricity price for a given hour in a day $X$ is equal to that of the same hour of day $X-1$.
    \item \textbf{Previous Week’s Price:} Assumes that the electricity price for a given hour in a day $X$ is equal to that of the same hour of day $X-7$.
    \item \textbf{Previous Month’s Price:} Assumes that the electricity price for a given hour in a day $X_{month}$ is equal to that of the same hour of day $X_{month-1}$.
\end{itemize}

As shown in \Cref{tab:baseline_models}, the Previous Day’s Price baseline achieves the lowest error among all baselines, with an MAE of 27.85 and an RMSE of 44.35. The Synthetic Price baseline performs the second best, but with a noticeably higher error. The Previous Week and Previous Month approaches yielded the highest MAE and RMSE values.

\begin{figure}[hbt!]
    \centering
    \includegraphics[width=1\linewidth]{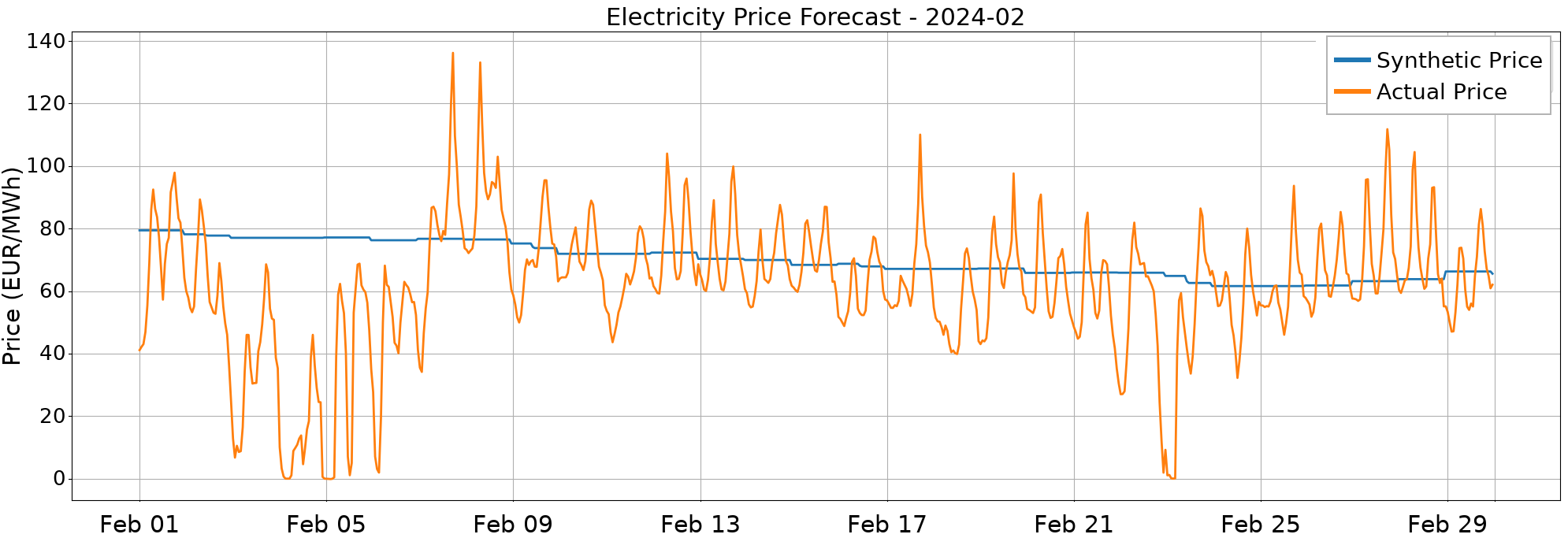}
    \caption{Comparison of actual prices (EUR/MWh) vs. the Synthetic Price baseline, which is derived from gas prices and CO\textsubscript{2} emission allowances.}
    \label{fig:synthetic_price}
\end{figure}

\begin{figure}[hbt!]
    \centering
    \begin{subfigure}{0.45\textwidth}
        \centering
        \includegraphics[width=\linewidth]{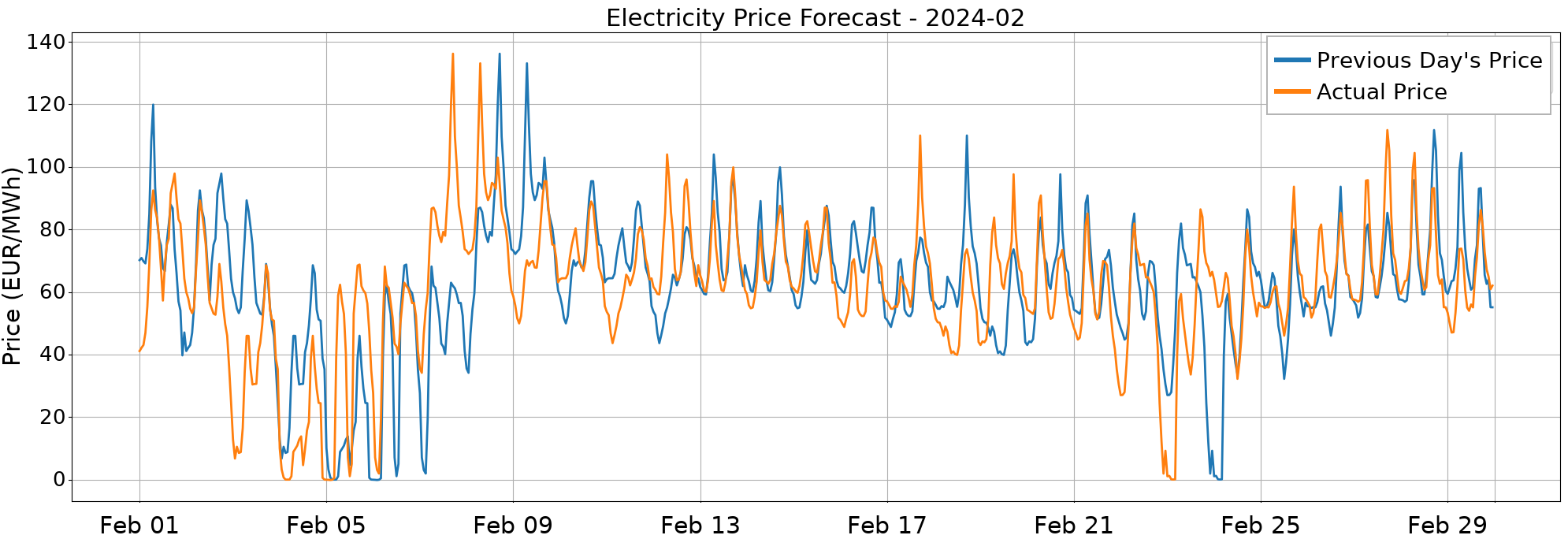}
        \caption{Previous Day's Price vs. Actual Price (EUR/MWh).}
        \label{fig:prev_day}
    \end{subfigure}
    \begin{subfigure}{0.45\textwidth}
        \centering
        \includegraphics[width=\linewidth]{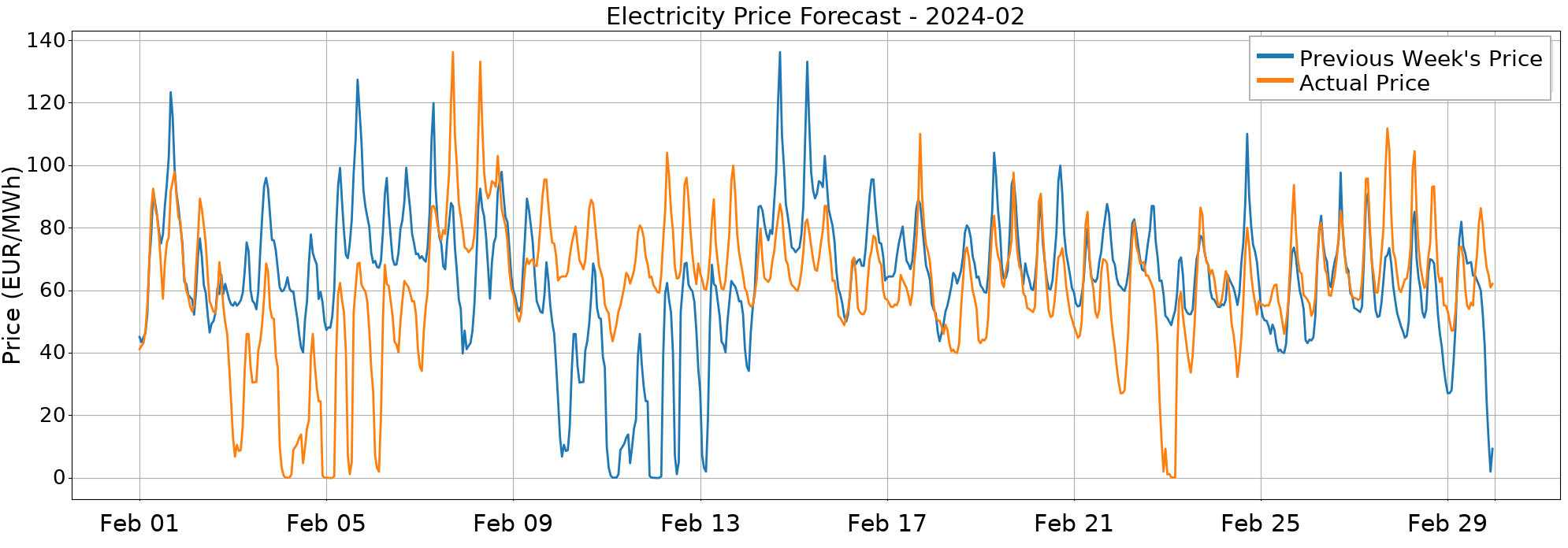}
        \caption{Previous Week's Price vs. Actual Price (EUR/MWh).}
        \label{fig:prev_week}
    \end{subfigure}
    \begin{subfigure}{0.45\textwidth}
        \centering
        \includegraphics[width=\linewidth]{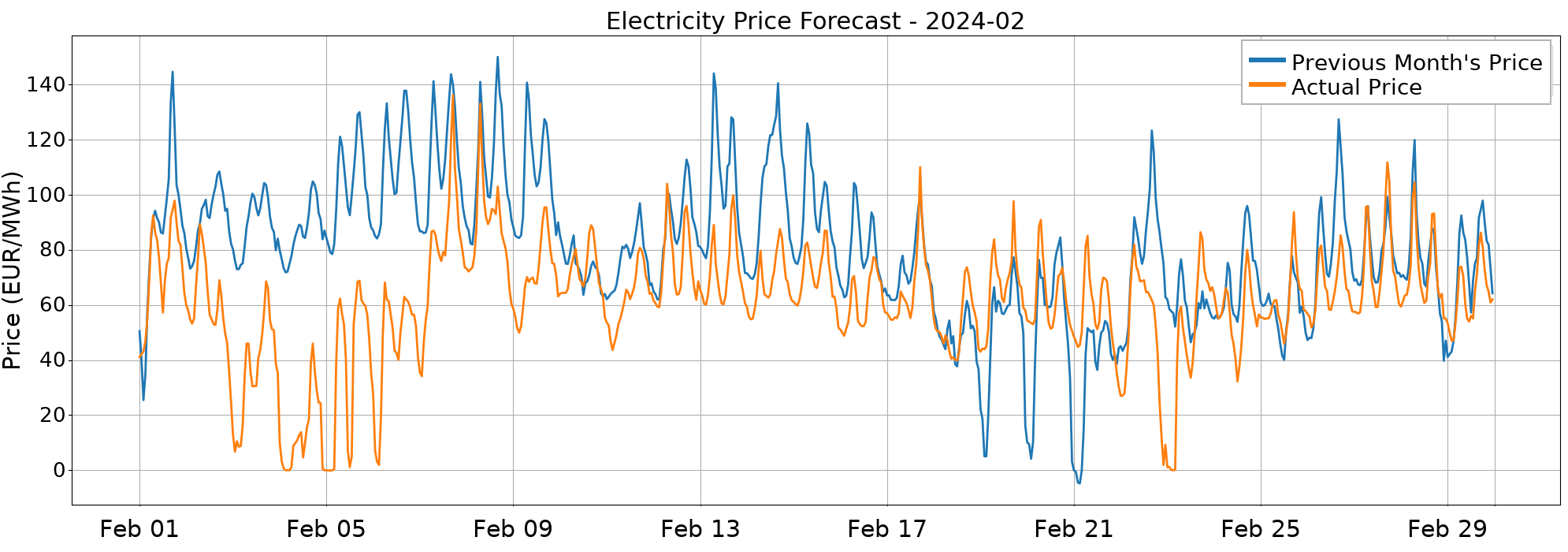}
        \caption{Previous Month's Price vs. Actual Price (EUR/MWh).}
        \label{fig:prev_month}
    \end{subfigure}
    \caption{Comparison of different baseline models (blue) against actual electricity prices (orange) for 2024 in EUR/MWh. The Previous Day’s Price performs best in stable periods.}
    \label{fig:baseline_models_comparison}
\end{figure}

 \Cref{fig:synthetic_price} and \Cref{fig:baseline_models_comparison} show how each baseline tracks the actual electricity price over 2024. The Previous Day's Price baseline closely follows the actual prices in stable periods, but misses sudden spikes. The Synthetic Price baseline computed from fuel costs remains relatively flat and cannot capture short-term volatility, although it offers a reasonable approximation of target prices.

Given its superior performance, we adopt the Previous Day’s Price as the primary reference benchmark for evaluating deep learning models in the following sections.

\begin{table}[hbt!]
\caption{Evaluation of Simple Baseline Models using MAE and RMSE. The best-performing model for each metric is highlighted in \textcolor{cyan}{\textbf{blue}}.}
\centering
\footnotesize
\rowcolors{2}{lightgray}{white}
\renewcommand{\arraystretch}{1.2} 
\begin{tabular}{lcc} 
\toprule
\tablehead{Baseline Model} & \tablehead{MAE} & \tablehead{RMSE} \\
\midrule
\textbf{Previous Day’s Price}  & \cellcolor{highlightblue}\textbf{27.85}  & \cellcolor{highlightblue}\textbf{44.35}  \\
\textbf{Synthetic Price}       & 31.97  & 51.49  \\
\textbf{Previous Week’s Price} & 32.40  & 55.11  \\
\textbf{Previous Month’s Price} & 33.01  & 51.22  \\
\bottomrule
\end{tabular}
\label{tab:baseline_models}
\end{table}

\subsubsection{Deep learning models}
\label{sec:deep_learning_models}

The models evaluated in this section differ in architectural design, training dataset size, and scope of the feature input. We begin with default implementations provided by the NeuralForecast library, applying minimal hyperparameter tuning. For each model, we additionally compared the three candidate training windows introduced in \Cref{sec:methods} (starting in 2018, 2020, and 2023), together with different train--validation split choices, and report only the best-performing configuration per model in \Cref{tab:model-performance-calendar}. As a result, the reported start year varies across models: it reflects a per-model selection of the most favorable training window rather than a single fixed training period. Because only the best configuration per model is shown, absolute errors should be compared across architectures with this selection in mind.

Overall, the deep-learning models consistently outperform the baseline methods described in \Cref{sec:baseline_models}. Among them, TFT achieves the lowest MAE and RMSE when calendar features are included, whereas NBEATSx achieves the lowest MAE and RMSE when using only electricity
 prices. Both models benefit from calendar information, suggesting that temporal structures such as weekdays and holidays may serve as effective predictive signals.

Despite its widespread use, LSTM shows degraded performance when calendar features are added, which may reflect overfitting or suboptimal handling of categorical time features in this architecture. Similarly, VT, which is capable of modeling long-range dependencies, yields the weakest overall performance, potentially owing to limited inductive bias or insufficient regularization.

Mamba, a recent state-space model, demonstrates improved performance when calendar features are included, reducing the MAE from 27.45 to 22.72. This suggests that the model can benefit from structured temporal inputs even with a relatively simple design.

\Cref{tab:model-performance-calendar} summarizes these findings. All results are reported for the same test period and input window (7 days); however, because the training start year was tuned per model, comparisons of absolute error across architectures should be interpreted with some caution.

\begin{table*}[htbp!]
  \caption{Model performance with and without calendar features (EP = electricity price only; Cal = calendar features). Best value per column is bold.}
  \label{tab:model-performance-calendar}
  \centering
  \begin{tabular}{lcccccc}
    \toprule
    Model & Start year & MAE & MAE & RMSE & RMSE & Input days \\
     &  & (EP) & (EP+Cal) & (EP) & (EP+Cal) & \\
    \midrule
    Mamba & 2018 & 27.45 & 22.72 & 44.36 & 39.31 & 7 \\
    NBEATSx & 2023 & \textbf{22.32} & 22.10 & \textbf{37.34} & 37.48 & 7 \\
    N-HiTS & 2020 & 23.8 & 21.9 & 45.02 & 37.42 & 7 \\
    LSTM & 2023 & 22.92 & 24.88 & 38.77 & 43.70 & 7 \\
    TFT & 2018 & 23.14 & \textbf{21.69} & 40.22 & \textbf{36.13} & 7 \\
    VT & 2018 & 32.42 & 26.91 & 50.84 & 43.74 & 7 \\
    \bottomrule
  \end{tabular}
\end{table*}

\subsection{Exogenous feature ablation}
\label{sec:exogenous_ablation}

To examine the contribution of exogenous inputs, we conduct a mini-ablation study, as summarized in \Cref{tab:ablation-results}. All models use a fixed 7-day input window, which yields the most favorable results. We report only the best-performing configuration per model under different input combinations.

The columns in \Cref{tab:ablation-results} indicate whether the gas prices, CO\textsubscript{2} allowances, synthetic prices, and loads are included in the input. Calendar features and electricity prices are always included. The MAE and RMSE values under the model's name reflect earlier benchmarks using electricity prices and calendar features alone, whereas the corresponding columns show performance with added exogenous features.

Several trends are observed. For many models, gas and synthetic price features offer modest gains when selectively used. For example, NBEATSx achieves the best MAE (20.90) with gas only, while adding more features leads to diminished returns. N-HiTS performs the best (MAE 20.20, RMSE 35.11) with gas, CO\textsubscript{2}, and synthetic prices, but without load.

The TFT model exhibits an inconsistent performance when additional features are included. The lowest MAE and RMSE are achieved using gas and synthetic price inputs; however, these results are still worse than those obtained using only electricity prices and calendar features.

Similarly, LSTM performs best with gas and calendar inputs. Including additional features increases the error, suggesting a sensitivity to noisy inputs. The best performance is obtained when limited to electricity prices and calendar features.

VT performs poorly across all configurations, reinforcing earlier observations of weak performance, possibly owing to the lack of extensive optimization at this step.

Interestingly, the Mamba model achieves a slight gain when trained with only gas prices (MAE = 22.51), outperforming its calendar-only counterpart. However, the addition of further exogenous features degrades its accuracy, suggesting that Mamba could be better suited to clean, low-dimensional inputs.

Overall, this study indicates that exogenous inputs can be helpful, but more features do not guarantee better performance.

\begin{table}
  \caption{Mini ablation study of exogenous features by model. Each row shows a different input combination. Best MAE and RMSE for each model are shown in bold. Under each model's name is its previous best MAE and RMSE value.}
  \label{tab:ablation-results}
  \centering
  \footnotesize
  \resizebox{\columnwidth}{!}{%
  \begin{tabular}{lcccccccc}
    \toprule
    Model &Year&MAE&RMSE&Gas&Synth&CO\textsubscript{2}&Load\\
    \midrule

    \textbf{Mamba} &&&&&&& \\
    MAE: 22.72 & 2018 & \textbf{22.51} & \textbf{38.89} & \cmark &  &  &  \\
    RMSE: 39.31 & 2018 & 26.84 & 44.94 & \cmark & \cmark &  &  \\
    
    \midrule
    \textbf{NBEATSx} &&&&&&& \\
    MAE: 22.10  & 2023 & \textbf{20.90} & \textbf{35.81} & \cmark &  &  &  \\
    RMSE: 37.34 & 2023 & 20.99 & 36.27 & \cmark & \cmark &  &  \\
               & 2023 & 26.58 & 44.64 & \cmark & \cmark & \cmark & \cmark  \\
    
    \midrule
    \textbf{N-HiTS} &&&&&&& \\
    MAE: 21.90  & 2020 & 20.74 & 36.61 & \cmark &  &  &  \\
    RMSE: 37.42 & 2020 & 20.65 & 35.73 & \cmark & \cmark &  &  \\
               & 2020 & \textbf{20.20} & \textbf{35.11} & \cmark & \cmark & \cmark &  \\
               & 2020 & 22.29 & 39.62 & \cmark & \cmark & \cmark & \cmark  \\
    \midrule
    \textbf{LSTM} &&&&&&& \\
    MAE: 22.92  & 2023 & \textbf{25.30} & \textbf{42.99} & \cmark &  &  &  \\
    RMSE: 38.77 & 2023 & 25.69 & 43.28 & \cmark & \cmark &  &  \\
               & 2023 & 27.48 & 48.22 & \cmark & \cmark & \cmark & \cmark  \\
    
    \midrule
    \textbf{TFT} &&&&&&& \\
    MAE: 21.69  & 2018 & 23.46 & 39.02 & \cmark &  &  &  \\
    RMSE: 36.13 & 2018 & \textbf{22.37} & \textbf{38.02} & \cmark & \cmark &  &  \\
               & 2018 & 22.79 & 39.16 & \cmark & \cmark & \cmark & \cmark  \\
    
    \midrule
    \textbf{VT} &&&&&&& \\
    MAE: 27.98  & 2020 & \textbf{30.82} & \textbf{47.79} & \cmark &  &  &  \\
    RMSE: 44.94 &&&&&&& \\
    
    \bottomrule
  \end{tabular}
  }
\end{table}

\subsection{Hyperparameter optimization}
\label{sec:hyperparam_optimization}

We perform model-specific hyperparameter tuning to further enhance the forecasting performance. Each model is evaluated using combinations of electricity price and calendar features, with additional exogenous inputs tested as relevant. The best results are presented in \Cref{tab:sp-load-ablation}.

The best-performing configurations for NBEATSx, N-HiTS, and TFT include gas, CO\textsubscript{2}, and synthetic price, while omitting the load. This aligns with the ablation findings (\Cref{sec:exogenous_ablation}), suggesting that moderate feature sets may outperform more extensive ones. N-HiTS achieves the lowest MAE (18.75) and RMSE (34.77), which can be attributed to its hierarchical interpolation and decomposition.

For NBEATSx, the optimal training set begins in 2020, indicating that a longer historical window supports improved generalization. Similar trends are observed for N-HiTS and TFT, which perform best when trained from 2020 and 2018 respectively.

Mamba and LSTM are excluded from this round of tuning due to time constraints. Although Mamba shows encouraging performance earlier, it requires further tuning. The LSTM has weaker overall results and is therefore deprioritized.

\begin{table}
  \caption{Model performance with and without SP features and load during hyperparameter optimization. 
  SP features include gas price, CO\textsubscript{2} emission allowances, and synthetic price.
  Models without these use only electricity price and calendar features. Bold values indicate the lowest value per method if there are multiple values.}
  \label{tab:sp-load-ablation}
  \centering
  \footnotesize
  \begin{tabular}{lcccccc}
    \toprule
    Model & Year & MAE & RMSE & SP features & Load \\
    \midrule
    NBEATSx & 2020 & \textbf{19.11} & \textbf{34.93} & \cmark & \\
    NBEATSx & 2020 & 20.09 & 35.28 & \cmark & \cmark \\
    N-HiTS & 2020 & \textbf{18.75} & \textbf{34.77} & \cmark & \\
    N-HiTS & 2020 & 21.70 & 37.29 & \cmark & \cmark \\
    TFT & 2018 & \textbf{19.40} & \textbf{35.63} & \cmark & \\
    TFT & 2018 & 20.51 & 35.88 & \cmark & \cmark \\
    VT & 2018 & 22.17 & 37.22 & & \\
    \bottomrule
  \end{tabular}
\end{table}

\subsection{Cross-zone generalization: zero-shot, one-shot, and few-shot learning}
\label{sec:learning_strategies}

\begin{figure}[htbp!]
    \centering
    \includegraphics[width=1\linewidth]{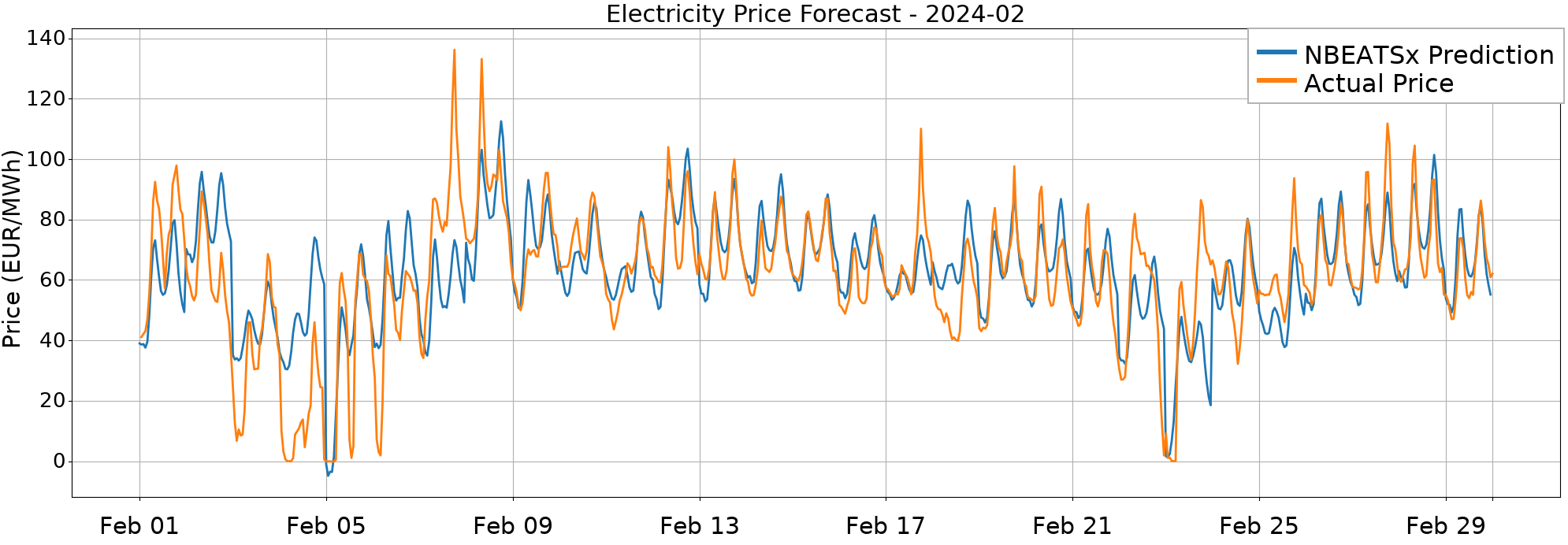}
    \caption{Illustration of a forecast using the NBEATSx model with One-Shot Learning (OSL). Achieving an MAE of 18.48 EUR/MWh, this model closely tracks price dynamics with only few exceptions. 
    The orange line represents the actual electricity price, while the blue line shows the NBEATSx’s prediction.}
    \label{fig:nbeatsx_best}
\end{figure}

To evaluate cross-market generalization, we conduct a series of learning strategy experiments focusing on transfer learning to the target zone of the DE-LU bidding zone. We first select the N-HiTS model's best-performing architecture after hyperparameter optimization as the basis for determining the potentially best training data range for this task. We evaluate three candidate training set starting dates: January 2015, October 2018 (post-DE-AT-LU split), and January 2020. Among these, using data from 2015 onward provides the best results and is adopted for the remainder of this section.

All models are trained using this full multi-zone dataset (excluding DE-LU) and tested on the DE-LU bidding zone during the 2024 period. We compare the three learning strategies by varying only the amount of DE-LU data included during training.

    
    

\Cref{tab:shot-comparison} presents the MAE results for the four deep learning models under these scenarios. Across all models and settings, the MAE falls within a narrow band (roughly 18.5--19.4), so the differences between models and between learning strategies are small. Because we do not report results across multiple seeds (see \Cref{sec:limitations}), part of this variation may reflect run-to-run variability rather than systematic effects, and the trends described below should be read as tentative. With this caveat in mind, N-HiTS appears to generalize well with minimal DE-LU exposure: its zero-shot and few-shot results are nearly identical (18.61 MAE) and its one-shot result is only marginally higher (18.65 MAE), which may indicate that very small amounts of local data are not sufficient to improve accuracy in this setting.

NBEATSx shows a slightly different pattern, reaching its lowest MAE in the one-shot setting (18.48), with a small increase in the few-shot setting (18.57); both improve modestly on its zero-shot result (18.96). This could indicate that a single example helps calibrate the model, although the magnitude of the effect is small and within the band noted above. The corresponding forecast is shown in \Cref{fig:nbeatsx_best}.

TFT performs best under zero-shot (18.63) and is somewhat higher under one-shot (19.35) and few-shot (19.13), which may suggest a preference for larger, more consistent context windows; here too the differences are modest.

VT achieves its best result in zero-shot (18.50) and is slightly higher under one-shot (18.77) and few-shot (18.96). More notably, VT improves substantially over its default configuration (MAE = 26.91; see \Cref{tab:model-performance-calendar}) once hyperparameter optimization and cross-zone transfer learning are applied. This suggests that much of the apparent gap observed for VT in earlier sections may stem from tuning budget rather than from the architecture itself.

Taken together, these results tentatively suggest that the four architectures perform comparably under low-data cross-zone conditions, and that their sensitivity to the amount of target-market data is modest. Confirming these trends would require repeated runs and statistical significance testing, which we leave to future work.

\begin{table}
  \caption{MAE results for zero-shot, one-shot, and few-shot learning across models. Bold values here indicate lowest value per strategy or column.}
  \label{tab:shot-comparison}
  \centering
  \footnotesize
  \begin{tabular}{lccc}
    \toprule
    Model & Zero-shot & One-shot & Few-shot \\
    \midrule
    N-HiTS   & 18.61 & 18.65 & 18.61 \\
    NBEATSx & 18.96 & \textbf{18.48} & \textbf{18.57} \\
    TFT     & 18.63 & 19.35 & 19.13 \\
    VT      & \textbf{18.50} & 18.77 & 18.96 \\
    \bottomrule
  \end{tabular}
\end{table}

\section{Limitations and future work}
\label{sec:limitations}

This study provides a comparative evaluation of a few deep learning models for day-ahead electricity price forecasting (EPF); however, several limitations remain. These findings also suggest directions for future research.

\textbf{Hyperparameter optimization and feature selection}:
Although we apply model-specific hyperparameter tuning (via Optuna) for selected architectures, the search spaces are constrained in scope, and the number of trials is limited owing to time and resource constraints. Future work could expand the search to include a broader set of hyperparameters, particularly for models with a large capacity and sensitivity to initialization (e.g., TFT and VT). Although we conduct a mini-ablation of exogenous inputs, this analysis is not exhaustive. A more comprehensive approach could jointly optimize both feature subsets and model configurations, which may better reveal model-feature interactions and improve the overall reliability.

\textbf{Statistical robustness and reporting}:
This study relies on MAE and RMSE, which are widely used but do not fully capture operationally relevant forecasting behavior. Future evaluations may benefit from additional metrics, such as the relative mean absolute error (rMAE) or symmetric mean absolute percentage error (sMAPE), and cost-sensitive or directional metrics. Additionally, our analysis primarily reports the best-performing configurations (features, dataset, etc.) as a summary of the experimental results, which could provide an incomplete view of model performance. Due to computational constraints, we do not report results across multiple seeds for every experiment; consequently, the reported outcomes may be subject to initialization bias. Furthermore, we do not include confidence intervals, error bars, or statistical significance tests in our analysis. Although a consistent pipeline is used across models, sources of variability--such as random initialization and training data splits--are not formally quantified. Furthermore, applying statistical tests, such as the Giacomini–White test \cite{giacomini_tests_2006}, would provide a more robust basis for model comparison.

\textbf{Computational documentation}:
Experiments are conducted using the BwUniCluster Jupyter interface (\Cref{sec:computation}). While training times and hardware details are recorded for the final model runs, we do not log resource usage (e.g., GPU memory), batch sizes, or training time per trial (and by extension, $CO\textsubscript{2}$ footprint) systematically. Future studies could benefit from logging more detailed computational metadata, which would aid reproducibility and facilitate more meaningful comparisons of the training efficiency and scalability across architectures.

\textbf{Scope of modeling contributions}:
This study does not introduce a novel machine learning architecture. While this may be seen as a limitation, our objective is to construct a relevant and realistic framework and systematically compare existing methods before proposing new ones. In our view, establishing a reproducible and well-documented baseline is a prerequisite for more targeted architectural innovation. We hope that this foundation will inform the development of new models that are better aligned with the specific challenges of EPF.

\textbf{Next steps}:
Building on this work and addressing the limitations discussed above,  possible extensions include benchmarking probabilistic forecasting methods, such as quantile regression or distributional models; evaluating the transferability of foundation models in the EPF setting; and incorporating interpretability tools, such as SHAP, attention attribution, or counterfactual explanations. These directions may improve transparency and decision support, especially in high-stakes applications.

\section{Conclusion}
\label{sec:sum}

We present a comparative study of deep learning models for day-ahead EPF with a focus on cross-border generalization. Our evaluation covered six architectures and three learning strategies: zero-shot, one-shot, and few-shot using consistent test data from the DE-LU bidding zone in 2024.

The results suggest that NBEATSx and N-HiTS perform reliably across multiple scenarios, particularly when calendar features and selected exogenous variables are used, although the differences between the strongest models are often small and would benefit from confirmation across repeated runs. Transformer-based models also benefit from transfer learning, but tend to require more extensive tuning and longer training histories. Mamba, a lightweight state-space model, showed competitive results with minimal input features, highlighting its potential in data-scarce settings.

Our analysis suggests that the design of the input feature set and temporal alignment can be as influential as the model architecture. In many cases, selective incorporation of exogenous variables appears more effective than using broad or complex inputs. We also observe that small amounts of target-market data can improve performance under low-data conditions, although the benefit is model-dependent.

This study aims to support the development of reproducible benchmarks for EPF and to provide insight into how deep learning models generalize across markets and data regimes. We hope to contribute to ongoing efforts in market modeling and forecasting by framing the task within realistic operational constraints and cross-border learning settings. To facilitate future research, we leverage a publicly available dataset and provide standardized training procedures based on existing implementations of several deep learning methods. These models serve as baselines for fair and consistent comparisons with future approaches. Improved forecasting tools may assist in integrating renewable energy sources more effectively, which, in turn, supports broader goals related to energy system efficiency and climate change mitigation.

\appendices

\section{Data Licensing and Terms of Use}
\label{sec:license}
The data provided by the Energy-Charts API was licensed under the CC BY 4.0 license. The proper attribution of Energy-Charts.info as the source is required. 
The Fraunhofer-Gesellschaft retains full copyright over the content provided on the Energy Charts website. The downloading or printing of publications is permitted for personal use and for the purpose of reporting on the Fraunhofer-Gesellschaft and its institutes, in accordance with specified usage conditions. Any further use, especially commercial utilization and distribution, generally requires written permission \cite{burger_impressum_nodate}. Requests should be directed to 
Fraunhofer-Institut fuer Solare Energiesysteme ISE
Heidenhofstr. 2
79110 Freiburg
For full details on the terms of use and licensing, please refer to the official \href{https://www.energy-charts.info/publishing-notes.html?l=en&c=DE}{publishing notes} and \href{https://api.energy-charts.info/}{API licensing}.


\section{\break Computational Resources}
\label{sec:computation}
Although predictive accuracy is a critical factor in evaluating forecasting models, computational efficiency plays an equally important role, especially for real-world deployment. A model that marginally improves forecasting accuracy but requires significantly more computational resources may not be suitable for operational use.

\Cref{tab:computation_time} presents an overview of the training times of the tested models. These values reflect the time required for training a model on the full dataset and do not account for hyperparameter optimization, which typically requires multiple iterations and increases the overall training duration.

Beyond standard training, hyperparameter optimization introduces a significant computational burden, as it involves searching through multiple configurations to find the best-performing model. For all the tested models, the optimization process took between 3 and 4 hours, depending on the number of iterations and model complexity.
All experiments were conducted on the BwUniCluster \cite{noauthor_bwunicluster20jupyter_nodate},
which provides access to nodes equipped with NVIDIA A100 and H100 GPUs. The models were trained using the Jupyter notebook interface available on the cluster for interactive experimentation.

\begin{table}[hbt!]
\caption{Comparison of model training times. Times are provided as approximate ranges and measured on a GPU-enabled system.}
\centering
\footnotesize
\rowcolors{2}{lightgray}{white}
\renewcommand{\arraystretch}{1.2}
\begin{tabular}{lc}
\toprule
\tablehead{Model} & \tablehead{Training Time (minutes)} \\
\midrule
Mamba & 10--25 \\
LSTM & 15--30 \\
N-HiTS, NBEATSx & 20--40 \\
VT, TFT & 40--50 \\
\bottomrule
\end{tabular}
\label{tab:computation_time}
\end{table}

\textbf{Code and Data Availability:} The preprocessing pipeline, training code, and configuration files used in this study are available at \url{https://github.com/SaiShadow/Cross-Border-Electricity-Price-Forecasting-with-Deep-Learning}. The electricity market data is obtained from the Energy-Charts API under the terms described in \Cref{sec:license}.

\section*{Acknowledgment}
We gratefully acknowledge funding from the Helmholtz Association and Networking Fund through Helmholtz AI under Grant No. VH-NG-1727.

This work was performed on the computational resource bwUniCluster funded by the Ministry of Science, Research, and the Arts Baden-Württemberg and the Universities of the State of Baden-Württemberg, Germany, within the framework program bwHPC.

This paper was written during a time of rapid advancements in generative AI. While these tools streamlined tasks and improved efficiency, their outputs were always critically examined with careful consideration of their limitations and benefits. Rather than being a substitute for independent reasoning, they aid in supporting thoughtful analysis and judgment.
Specifically, the following AI tools were incorporated:
\begin{itemize}
  \item For experimental work, ChatGPT-4o, and more recently o3-mini and o3-mini-high, were used extensively to improve code readability, debugging, UI   prototyping, improving visualization, and refining experimental methodologies.
  \item For paper writing, ChatGPT-5 helped summarize parts of the text, improve structural coherence, refine table formatting, identify appropriate synonyms, improve overall text quality, and troubleshoot LaTeX formatting.
\end{itemize}
However, in order to ensure the originality and accuracy of the final content, human oversight continued to be central and necessary.

\bibliographystyle{IEEEtran}
\bibliography{ref}


\begin{IEEEbiographynophoto}{Hadeer Elashhab} received her B.S. degree in computer science from the American University in Cairo, Cairo, Egypt, in 2017
and her M.S. degree in artificial intelligence from Queen Mary University of London, London, United Kingdom, in 2022. She is currently pursuing her Ph.D.
degree in computer science at Karlsruhe Institute of Technology, Karlsruhe, Germany.

From 2017 to 2020, she was a Data Scientist with Raisa Energy LLC, Cairo, Egypt. Her research interest includes the
application of machine learning and deep learning 
techniques to energy time series, explanation of applied machine and deep learning methods, and the evaluation of explanation quality produced by such methods.

\end{IEEEbiographynophoto}

\begin{IEEEbiographynophoto}{Sai Srijan Papineni} received the B.Sc. degree in computer science from the Karlsruhe Institute of Technology (KIT), Karlsruhe, Germany, in 2025.

He is currently a Developer at SAP SE, Walldorf, Germany. Previously, he worked as a Working Student at SAP SE during his undergraduate studies.

Mr. Papineni’s research interests include deep learning–based time series forecasting, electricity price modeling, and the application of machine learning methods to energy markets.

\end{IEEEbiographynophoto}
\begin{IEEEbiography}[{\includegraphics[width=1in,height=1.25in,clip,keepaspectratio]{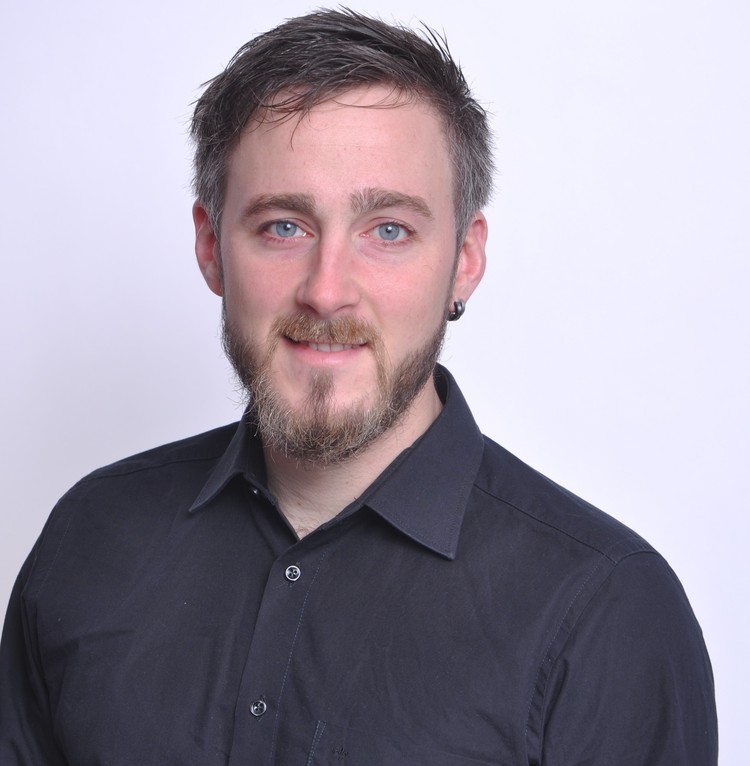}}]{Marvin Dorn} completed his Bachelor of Engineering in Electrical Engineering with a specialization in Automation at Karlsruhe University of Applied Sciences in 2019. Following his bachelor’s degree, he gained practical experience at Fraunhofer ICT, working on PEM fuel cell technologies. He then pursued a Master’s degree in Electrical Engineering at the Karlsruhe Institute of Technology (KIT), graduating in 2023 with a focus on Renewable Energies and a specialization in Electrochemical Systems, including batteries, fuel cells, and electrolysis. Since 2023, Marvin has been a Research Associate at the Institute for Automation and Applied Informatics (IAI) at KIT, where his research focuses on large-scale electrochemical systems and their integration into the electrical grid.
\end{IEEEbiography}

\begin{IEEEbiography}[{\includegraphics[width=1in,height=1.25in,clip,keepaspectratio]{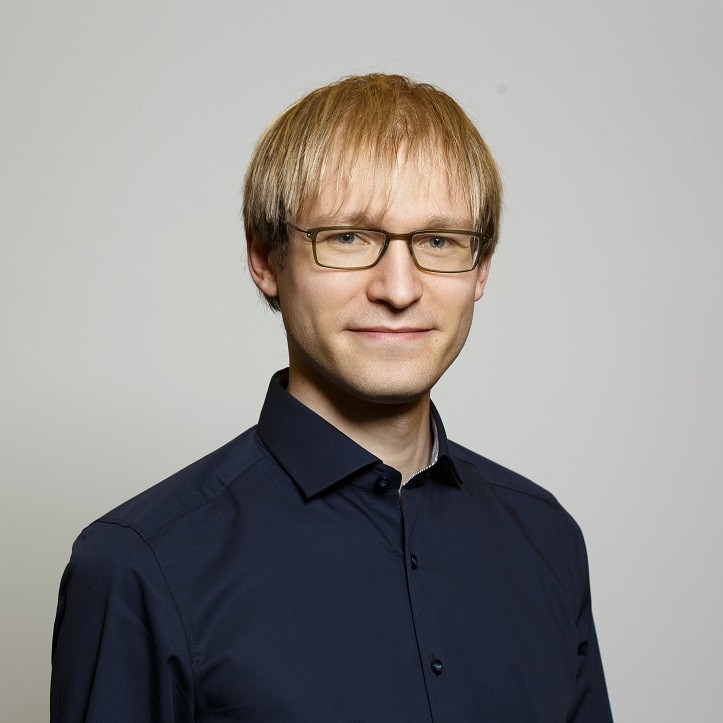}}]{Benjamin Schäfer} received his Diplom degree in Physics from the University of Magdeburg, Germany in 2013. Pursuing his Ph.D. in Göttingen (Germany) London (United Kingdom) and Tokyo (Japan), he received his Ph.D. degree in physics in 2017 from the University of Göttingen. He has worked as a Postdoctoral Researcher at the Max Planck Institute for Dynamics and Self-Organization, Göttingen, Germany, the Technical University Dresden, Germany and as a Marie Sklodowska-Curie Research Fellow at Queen Mary University of London. In 2021 he worked as Associate Professor in Physics at the Norwegian University of Life Sciences. Since 2022, he is leading a Young Investigator Group on Data-Driven Analysis of Complex Systems at the KIT, Germany, where he also holds an Assistant Professorship since 2023. He investigates power systems with the help of (explainable) machine learning and stochastic modeling.
\end{IEEEbiography}

\begin{IEEEbiographynophoto}{Veit Hagenmeyer}
Veit Hagenmeyer received a diploma in technical kybernetics from the University of Stuttgart, Stuttgart, Germany, in 1998, and a Docteur en Science in engineering from Universit\'e Paris, Paris, France, in 2002. His primary research interests are modeling, optimization, and control of sector-integrated energy systems, machine-learning-based forecasting for renewable-driven energy networks, and integrated cyber-security of such systems.

He is currently Professor of Energy Informatics with the Faculty of Informatics and Director of the Institute for Automation and Applied Informatics at the Karlsruhe Institute of Technology (KIT), Karlsruhe, Germany. From 2010 to 2014, he led three BASF power-plant sites and their grids (approximately 1~GW, about 200 staff, and a 750~million Euros annual budget). Earlier he served as Personal Assistant to the European Location Manager of BASF (2009--2010), Leader of the Advanced Process Control Group at BASF (2007--2008), Research Engineer in the same group (2003--2006), researcher on nonlinear control theory at the Institute of System Dynamics, University of Stuttgart (2003), and Postdoctoral Fellow on nonlinear control theory at L2S, C.N.R.S.-Sup\'elec-Universit\'e, Paris (2002). His current research continues to focus on integrated energy-system modeling, data-driven forecasting, and cyber-security.

Prof.\ Hagenmeyer is a member of the ACM e-Energy steering committee (2025) and holds several recognitions, including the ACM e-Energy Most Accepted Author award (2024), Best Reviewer award (2023), multiple KIT Faculty Awards for best teaching (2021, 2020, 2019, 2018, 2016), the BASF Innovation Awards Finalist (2009), the Lindau Meetings of Nobel Laureates Participation Award (2006), and the VDI Eugen-Hartmann Award (2005).
\end{IEEEbiographynophoto}

\EOD

\end{document}